# THE ISOCONDITIONING LOCI OF PLANAR THREE-DOF PARALLEL MANIPULATORS

**Damien Chablat**
Institut de Recherche en Communications et Cybernétique de Nantes[*], 1 rue de la Noë, BP 92101, 44321 Nantes Cedex 03, France, Tel : (33) 2 40 37 69 54, Fax : (33) 2 40 37 69 30
Damien.Chablat@irccyn.ec-nantes.fr

**Stéphane Caro**
Institut de Recherche en Communications et Cybernétique de Nantes[*], 1 rue de la Noë, BP 92101, 44321 Nantes Cedex 03, France, Tel : (33) 2 40 37 69 52, Fax : (33) 2 40 37 69 30
Stephane.Caro@irccyn.ec-nantes.fr

**Philippe Wenger**
Institut de Recherche en Communications et Cybernétique de Nantes[*], 1 rue de la Noë, BP 92101, 44321 Nantes Cedex 03, France, Tel : (33) 2 40 37 69 47, Fax : (33) 2 40 37 69 30
Philippe.Wenger@irccyn.ec-nantes.fr

**Jorge Angeles**
Centre for Intelligent Machines, McGill University, 817 Sherbrooke Street West, Montreal, Canada H3A 2K6, Tel : (514) 398-6315 , Fax : (514) 398-7348
Angeles@cim.mcgill.ca

**Abstract:**

*The subject of this paper is a special class of parallel manipulators. First, we analyze a family of three-degree-of-freedom manipulators. Two Jacobian matrices appear in the kinematic relations between the joint-rate and the Cartesian-velocity vectors, which are called the "inverse kinematics" and the "direct kinematics" matrices. The singular configurations of these matrices are studied. The isotropic configurations are then studied based on the characteristic length of this manipulator. The isoconditioning loci of all Jacobian matrices are computed to define a global performance index to compare the different working modes.*

## 1    Introduction

Various performance indices have been devised to assess the kinetostatic performances of serial and parallel manipulators. The literature on performance indices is extremely rich to fit in the limits of this paper, the interested reader are invited to look at it in the rather recent references cited here. A dimensionless quality index was recently introduced by Lee, Duffy, and Hunt [1] based on the ratio of the Jacobian determinant to its maximum absolute value, as applicable to parallel manipulators. This index does not take into account the location of the operation point of the end-effector, because the Jacobian determinant is independent of this location. The proof of the foregoing result is available in [2], as pertaining to serial manipulators, its extension to their parallel counterparts being straightforward. The *condition*

---

[*] IRCCyN : U.M.R. 6597, École Centrale de Nantes, École des Mines de Nantes, Université de Nantes.

*number* of a given matrix, on the other hand, is well known to provide a measure of invertibility of the matrix [3]. It is thus natural that this concept found its way in this context. Indeed, the condition number of the Jacobian matrix was proposed by Salisbury and Craig [4] as a figure of merit to minimize when designing manipulators for maximum accuracy. In fact, the condition number gives, for a square matrix, a measure of the relative roundoff-error amplification of the computed results [3] with respect to the data roundoff error. As is well known, however, the dimensional inhomogeneity of the entries of the Jacobian matrix prevents the straightforward application of the condition number as a measure of Jacobian invertibility. The *characteristic length* was introduced in [5] to cope with the above-mentioned inhomogeneity.

In this paper, we use the *characteristic length* to normalize the Jacobian matrix of a three-degree-of-freedom (dof) planar manipulator and to calculate the isoconditioning loci for all its working modes.

## 2   Preliminaries

A planar three-dof manipulator, with three parallel PRR chains, the object of this paper, is shown in Fig. 1. This manipulator has been frequently studied, in particular in [6, 7]. The actuated joint variables are the displacements of the three prismatic joints. The Cartesian variables are the position $P$ and the orientation $\theta$ of the platform.

The points $A_1$, $A_2$ and $A_3$ are located in the vertices of an equilateral triangle whose geometric center is the point $O$, as well as the points $C_1$, $C_2$ and $C_3$, whose geometric center is the point $P$. Point $P$, moreover, is the *operation point* of the manipulator. To reduce the number of design variables, we set $R=R_1=R_2=R_3$, with $R_i$ denoting the lengths of $A_iO$, $l=l_1=l_2=l_3$ with $l_i$ denoting the lengths of $B_iC_i$ and $r=r_1=r_2=r_3$ with $r_i$ denoting the lengths of $C_iP$.

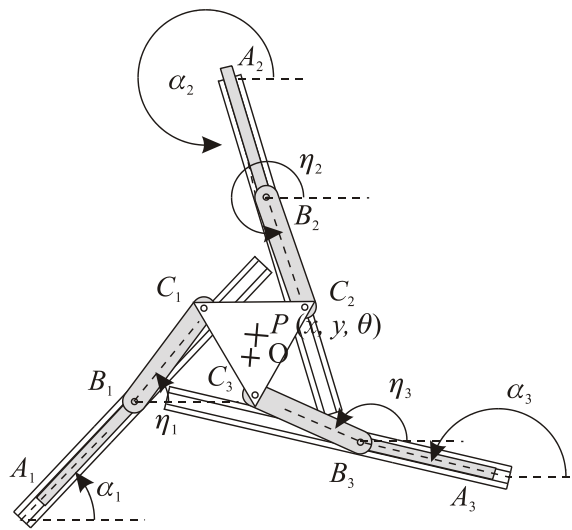

*Figure 1: A three-dof parallel manipulator*

## 2.1 Kinematic Relations

The velocity $\dot{\mathbf{p}}$ of the point *P*, of position vector **p**, can be obtained in three different forms, depending on the direction in which the loop is traversed, namely,

$$\dot{\mathbf{p}} = \dot{\mathbf{b}}_1 + \dot{\eta}_1 \mathbf{E} (\mathbf{c}_1 - \mathbf{b}_1) + \dot{\theta} \mathbf{E} (\mathbf{p} - \mathbf{c}_1) \tag{1a}$$

$$\dot{\mathbf{p}} = \dot{\mathbf{b}}_2 + \dot{\eta}_2 \mathbf{E} (\mathbf{c}_2 - \mathbf{b}_2) + \dot{\theta} \mathbf{E} (\mathbf{p} - \mathbf{c}_2) \tag{1b}$$

$$\dot{\mathbf{p}} = \dot{\mathbf{b}}_2 + \dot{\eta}_3 \mathbf{E} (\mathbf{c}_3 - \mathbf{b}_3) + \dot{\theta} \mathbf{E} (\mathbf{p} - \mathbf{c}_3) \tag{1c}$$

with matrix **E** defined as

$$\mathbf{E} = \begin{bmatrix} 0 & -1 \\ 1 & 0 \end{bmatrix}$$

The velocity $\dot{\mathbf{b}}_i$ of points $B_i$ is given by

$$\dot{\mathbf{b}}_i = \dot{\rho}_i \begin{bmatrix} \cos(\alpha_i) \\ \sin(\alpha_i) \end{bmatrix} = \dot{\rho}_i \, \boldsymbol{\alpha}_i$$

where $\boldsymbol{\alpha}_i$ are unit vectors, collinear with the prismatic joints.

We would like to eliminate the three idle joint rates $\dot{\eta}_1$, $\dot{\eta}_2$ and $\dot{\eta}_3$ from eqs.(1a-c), which we do upon dot-multiplying the former by $(\mathbf{c}_i - \mathbf{b}_i)^T$, thus obtaining

$$(\mathbf{c}_1 - \mathbf{b}_1)^T \dot{\mathbf{p}} = (\mathbf{c}_1 - \mathbf{b}_1)^T \dot{\mathbf{b}}_1 + (\mathbf{c}_1 - \mathbf{b}_1)^T \dot{\theta} \mathbf{E} (\mathbf{p} - \mathbf{c}_1) \tag{2a}$$

$$(\mathbf{c}_2 - \mathbf{b}_2)^T \dot{\mathbf{p}} = (\mathbf{c}_2 - \mathbf{b}_2)^T \dot{\mathbf{b}}_2 + (\mathbf{c}_2 - \mathbf{b}_2)^T \dot{\theta} \mathbf{E} (\mathbf{p} - \mathbf{c}_2) \tag{2b}$$

$$(\mathbf{c}_3 - \mathbf{b}_3)^T \dot{\mathbf{p}} = (\mathbf{c}_3 - \mathbf{b}_3)^T \dot{\mathbf{b}}_2 + (\mathbf{c}_3 - \mathbf{b}_3)^T \dot{\theta} \mathbf{E} (\mathbf{p} - \mathbf{c}_3) \tag{2c}$$

Equations (2a-c) can now be cast in vector form, namely,

$$\mathbf{A} \, \mathbf{t} = \mathbf{B} \, \dot{\boldsymbol{\rho}} \quad \text{with} \quad \mathbf{t} = \begin{bmatrix} \dot{\mathbf{p}} \\ \dot{\theta} \end{bmatrix}, \; \dot{\boldsymbol{\rho}} = \begin{bmatrix} \dot{\rho}_1 \\ \dot{\rho}_2 \\ \dot{\rho}_3 \end{bmatrix} \tag{3}$$

and $\dot{\boldsymbol{\rho}}$ defined as the vector of actuated joint rates. Moreover **A** and **B** are, respectively, the direct-kinematics and the inverse-kinematics matrices of the manipulator, defined as

$$\mathbf{A} = \begin{bmatrix} (\mathbf{c}_1-\mathbf{b}_1)^T & -(\mathbf{c}_1-\mathbf{b}_1)^T \mathbf{E} (\mathbf{p} - \mathbf{c}_1) \\ (\mathbf{c}_2-\mathbf{b}_2)^T & -(\mathbf{c}_2-\mathbf{b}_2)^T \mathbf{E} (\mathbf{p} - \mathbf{c}_2) \\ (\mathbf{c}_3-\mathbf{b}_3)^T & -(\mathbf{c}_3-\mathbf{b}_3)^T \mathbf{E} (\mathbf{p} - \mathbf{c}_3) \end{bmatrix} \tag{4a}$$

and

$$\mathbf{B} = \begin{bmatrix} (\mathbf{c}_1-\mathbf{b}_1)^T \boldsymbol{\alpha}_1 & 0 & 0 \\ 0 & (\mathbf{c}_2-\mathbf{b}_2)^T \boldsymbol{\alpha}_2 & 0 \\ 0 & 0 & (\mathbf{c}_3-\mathbf{b}_3)^T \boldsymbol{\alpha}_3 \end{bmatrix} \tag{4b}$$

When **A** and **B** are nonsingular, we obtain the relations

$$\mathbf{t} = \mathbf{J} \, \dot{\boldsymbol{\rho}} \; \text{ with } \; \mathbf{J} = \mathbf{A}^{-1} \mathbf{B} \quad \text{and} \quad \dot{\boldsymbol{\rho}} = \mathbf{K} \, \mathbf{t}, \text{ with } \mathbf{K} = \mathbf{B}^{-1} \mathbf{A}$$

## 2.2 Parallel Singularities

Parallel singularities occur when the determinant of matrix **A** vanishes [8, 9]. In this configuration, it is possible to move locally the operation point *P* with the actuators locked. These singularities are to be avoided, because the structure cannot resist arbitrary forces, and control is lost. To avoid any performance deterioration, it is necessary to have a Cartesian workspace free of parallel singularities. For the planar manipulator studied, such configurations are reached whenever the axes $B_1C_1$, $B_2C_2$ and $B_3C_3$ intersect, possibly at infinity, as depicted in Fig. 2.

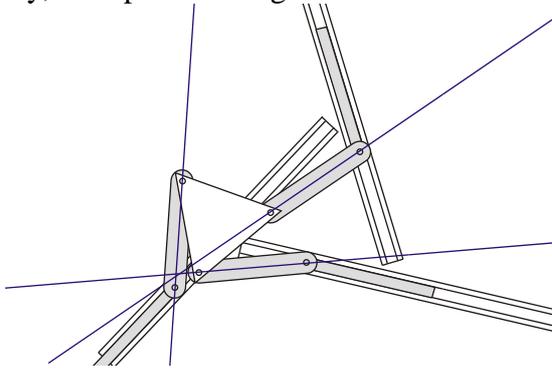 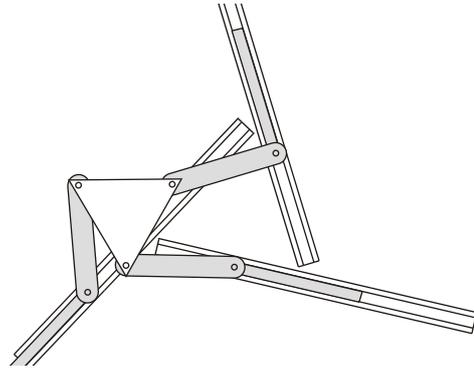

*Figure 2: Parallel singularity*  *Figure 3: Serial singularity*

In such configurations, the manipulator cannot resist a force applied at the intersection point. These singular configurations are located inside the Cartesian workspace and form the boundaries of the joint workspace [8].

## 2.3 Serial Singularities

Serial singularities occur when $\det(\mathbf{B}) = 0$. In the presence of theses singularities, there is a direction along which no Cartesian velocity can be produced. Serial singularities define the boundary of the Cartesian workspace. For the topology under study, the serial singularities occur whenever $A_1B_1 \perp B_1C_1$ or $A_2B_2 \perp B_2C_2$ or $A_3B_3 \perp B_3C_3$ as depicted in Fig. 3.

## 3 Isoconditioning Loci

### 3.1 The Matrix Condition Number

We derive below the loci of equal condition number of the direct-and inverse-kinematics matrices. To do this, we first recall the definition of *condition number* $\kappa(\mathbf{M})$ of an $m \times n$ matrix **M**, with $m \leq n$. This number can be defined in various ways; for our purposes, we define $\kappa(\mathbf{M})$ as the ratio of the largest, $\sigma_l$, to the smallest, $\sigma_s$, singular values of **M**, namely,

$$\kappa(\mathbf{M}) = \frac{\sigma_l}{\sigma_s} \tag{5}$$

The singular values $\{\sigma_i\}_1^m$ of matrix **M** are defined, in turn, as the square roots of the nonnegative eigenvalues of the positive definite $m \times m$ matrix $\mathbf{MM}^T$.

### 3.2 Non-Homogeneous Direct-Kinematics Matrix

It will prove convenient to partition **A** into a 2×3 block $\mathbf{A}_1$ and a 1×3 block $\mathbf{A}_2$, defined as

$$\mathbf{A}_1 = \begin{bmatrix} (\mathbf{c}_1\text{-}\mathbf{b}_1)^T \\ (\mathbf{c}_2\text{-}\mathbf{b}_2)^T \\ (\mathbf{c}_3\text{-}\mathbf{b}_3)^T \end{bmatrix} \text{ and } \mathbf{A}_2 = \begin{bmatrix} -(\mathbf{c}_1\text{-}\mathbf{b}_1)^T \mathbf{E}\,(\mathbf{p}-\mathbf{c}_1) \\ -(\mathbf{c}_2\text{-}\mathbf{b}_2)^T \mathbf{E}\,(\mathbf{p}-\mathbf{c}_2) \\ -(\mathbf{c}_3\text{-}\mathbf{b}_3)^T \mathbf{E}\,(\mathbf{p}-\mathbf{c}_3) \end{bmatrix} \quad (6)$$

Therefore, while the entries of $\mathbf{A}_1$ have units of length, those of $\mathbf{A}_2$ have units of square length. It is thus apparent that the singular values of **A** have different dimensions and hence, it is impossible to compute κ(**A**) because the singular values of **A** cannot be ordered from smallest to largest. The normalization of the Jacobian for purposes of rendering it dimensionless has been judged to be dependent on the normalizing length [10, 11]. As a means to avoid the arbitrariness of the choice of that normalizing length, the characteristic length *L* was introduced in [12].

To yield the matrix **A** homogeneous, each term of the third column of **A** is divided by the characteristic length *L*, thereby deriving its normalized counterpart $\bar{\mathbf{A}}$:

$$\bar{\mathbf{A}} = \begin{bmatrix} (\mathbf{c}_1\text{-}\mathbf{b}_1)^T & -(\mathbf{c}_1\text{-}\mathbf{b}_1)^T \mathbf{E}\,(\mathbf{p}-\mathbf{c}_1)/L \\ (\mathbf{c}_2\text{-}\mathbf{b}_2)^T & -(\mathbf{c}_2\text{-}\mathbf{b}_2)^T \mathbf{E}\,(\mathbf{p}-\mathbf{c}_2)/L \\ (\mathbf{c}_3\text{-}\mathbf{b}_3)^T & -(\mathbf{c}_3\text{-}\mathbf{b}_3)^T \mathbf{E}\,(\mathbf{p}-\mathbf{c}_3)/L \end{bmatrix} \quad (7)$$

which is calculated so as to minimize $\kappa(\bar{\mathbf{A}})$, along with the posture variables $\rho_1$, $\rho_2$ and $\rho_3$.

### 3.3 Isotropic Configuration

In this section, we write the isotropy condition in **J** to define the geometric parameters of the manipulator. We shall obtain also the value *L* of the characteristic length. To simplify **A** and **B**, we use the notation

$$\mathbf{l}_i = \mathbf{c}_i - \mathbf{b}_i \quad (8a)$$
$$k_i = (\mathbf{c}_i - \mathbf{b}_i)^T \mathbf{E}\,(\mathbf{p}-\mathbf{c}_i) \quad (8b)$$
$$m_i = (\mathbf{c}_i - \mathbf{b}_i)^T \boldsymbol{\alpha}_i \quad (8c)$$
$$\gamma_i = \angle B_i C_i P \quad (8d)$$

We can write matrices $\bar{\mathbf{A}}$, **B** and $\mathbf{B}^{-1}$ as,

$$\bar{\mathbf{A}} = \begin{bmatrix} \mathbf{l}_1^T & k_1/L \\ \mathbf{l}_2^T & k_2/L \\ \mathbf{l}_3^T & k_3/L \end{bmatrix},\; \mathbf{B} = \begin{bmatrix} m_1 & 0 & 0 \\ 0 & m_2 & 0 \\ 0 & 0 & m_3 \end{bmatrix} \text{ and } \mathbf{B}^{-1} = \begin{bmatrix} 1/m_1 & 0 & 0 \\ 0 & 1/m_2 & 0 \\ 0 & 0 & 1/m_3 \end{bmatrix}$$

Whenever matrix **B** is nonsingular, that is when $m_i \neq 0$, for $i=1,2,3$. By using the previous simplifications, we have

$$\bar{\mathbf{K}} = \begin{bmatrix} \mathbf{l}_1^T/m_1 & -k_1/(L\,m_1) \\ \mathbf{l}_2^T/m_2 & -k_2/(L\,m_2) \\ \mathbf{l}_3^T/m_3 & -k_3/(L\,m_3) \end{bmatrix}$$

Matrix $\bar{\mathbf{J}}$, the normalized **J**, is isotropic if and only if $\bar{\mathbf{K}}\bar{\mathbf{K}}^T = \tau^2 \mathbf{1}_{3\times 3}$ for $\tau > 0$, *i.e.*,

$$(\mathbf{l}_1^T \mathbf{l}_1 + k_1^2/L^2)/m_1^2 = \tau^2 \quad (9a)$$

$(\mathbf{l}_2^T\mathbf{l}_2 + k_2^2/L^2)/m_2^2 = \tau^2$ (9b)

$(\mathbf{l}_3^T\mathbf{l}_3 + k_3^2/L^2)/m_3^2 = \tau^2$ (9c)

$(\mathbf{l}_1^T\mathbf{l}_2 + k_1 k_2/L^2)/(m_1 m_2) = 0$ (9d)

$(\mathbf{l}_1^T\mathbf{l}_3 + k_1 k_3/L^2)/(m_1 m_3) = 0$ (9e)

$(\mathbf{l}_2^T\mathbf{l}_3 + k_2 k_3/L^2)/(m_2 m_3) = 0$ (9f)

From eqs. (9a-f), we can derive the conditions below:

$\|\mathbf{l}_1\| = \|\mathbf{l}_2\| = \|\mathbf{l}_3\|$ (10a)

$\|\mathbf{p}-\mathbf{c}_1\| = \|\mathbf{p}-\mathbf{c}_2\| = \|\mathbf{p}-\mathbf{c}_3\|$ (10b)

$\mathbf{l}_1^T\mathbf{l}_2 = \mathbf{l}_1^T\mathbf{l}_3 = \mathbf{l}_2^T\mathbf{l}_3$ (10c)

$m_1 m_2 = m_1 m_3 = m_2 m_3$ (10d)

In summary, the constraints defined in eqs.(10a-d) are
- Pivots $C_i$ should be placed in the vertices of an equilateral triangle, i.e. $r = r_1 = r_2 = r_3$;
- Segments $A_iB_i$ are the sides of an equilateral triangle;
- Segments $B_iC_i$ are the sides of an equilateral triangle;
- Segments $B_iC_i$ are equal, i.e. $l = l_1 = l_2 = l_3$.

We can notice that the first and the last conditions have already been proposed in § 2.

### 3.4 Characteristic Length

The characteristic length is defined at the isotropic configuration. From eqs.(9d-f), we determine the value of the characteristic length as,

$$L = \sqrt{\frac{-k_1 k_2}{\mathbf{l}_1^T\mathbf{l}_2}}$$

By applying the constraints defined in eqs.(10a-d), we can write the characteristic length in terms of angle $\gamma$, i.e.

$L = \sqrt{2}\, r \sin(\gamma)$

where $\gamma = \gamma_1 = \gamma_2 = \gamma_3$ are defined in eq. (8d) and $\gamma \in [0, 2\pi]$.

This means, that the manipulator under study possesses several isotropic configurations (Fig. 4a-b), whereas the characteristic length $L$ of the manipulator is unique [2]. When $\gamma$ is equal to $\pi/2$ (Fig. 4a), we obtain the condition:
- $B_iC_i$ is perpendicular to $C_iG$.

This condition yields a configuration furthest away from parallel singularities. To have a isotropic configuration furthest away from serial singularities, we have two conditions:
- $A_iB_i$ is collinear with $B_iC_i$;
- $r = R/2$.

Figure 4 depicts two isotropic configurations.

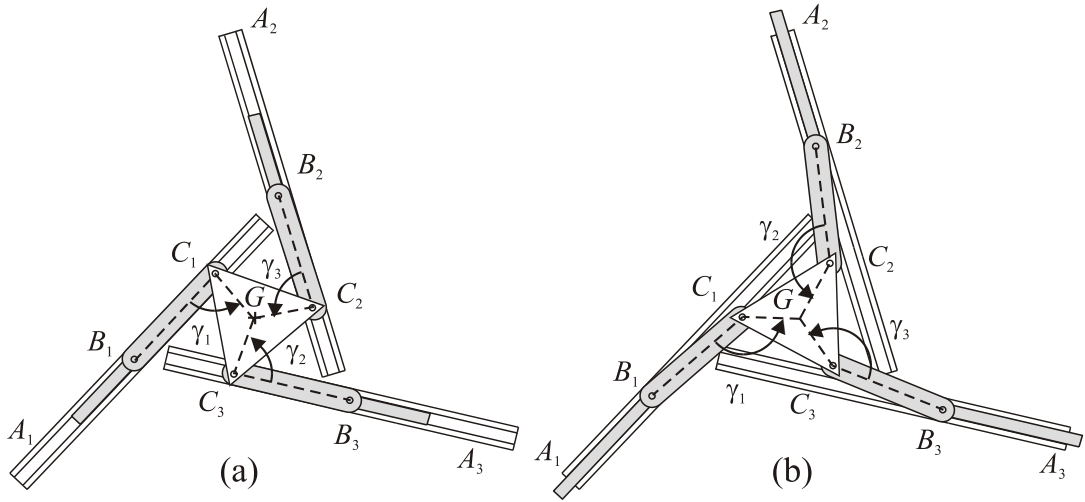

*Figure 4: Two isotropic configurations with two values of γ*

### 3.5 Working Modes

The manipulator under study has a diagonal inverse-kinematics matrix **B**, as shown in eq.(4b), the vanishing of one of its diagonal entries thus indicating the occurrence of a *serial singularity*. The set of manipulator postures free of this kind of singularity is termed a *working mode*. The different working modes are thus separated by a serial singularity, with a set of postures in different working modes corresponding to an inverse kinematics solution.

The formal definition of the working mode is detailed in [8]. For the manipulator at hand, there are eight working modes, as depicted in Fig. 5.

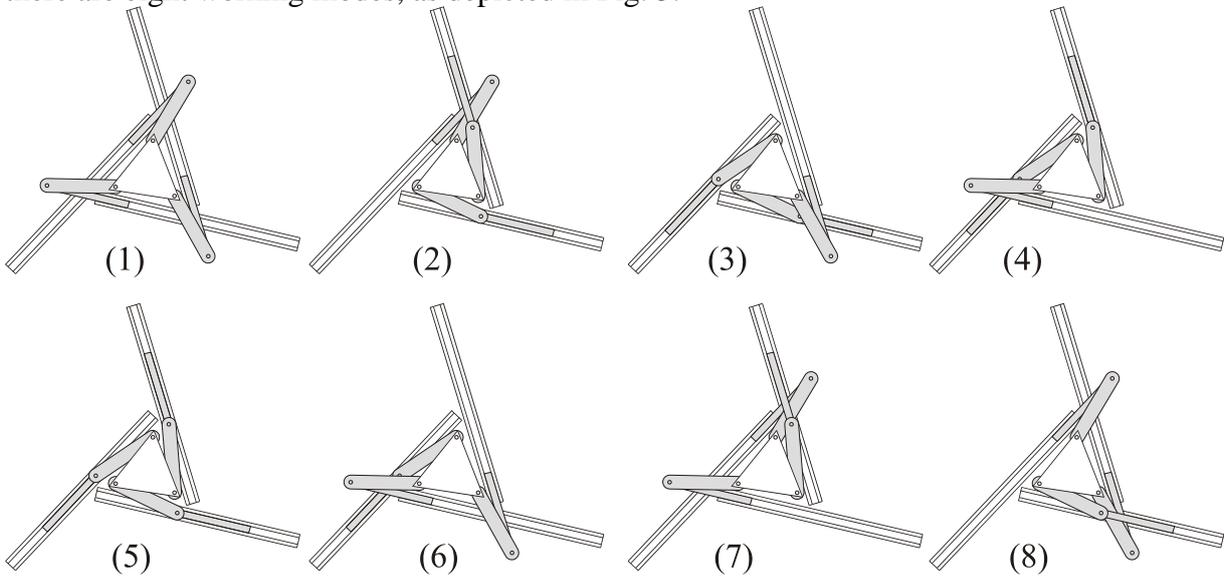

*Figure 5: The eight working modes of the 3PRR manipulator*

However, because of symmetry, we can restrict our study to only two working modes, if there are no joint limits. Indeed, the working mode 1 is similar to the working mode 5, because in a case the signs $m_i$ of the diagonal entries of **B** are either all positive, or all negative. We can associate the working modes 2-6, 3-7 and 4-8 likewise the working modes 3-4 and 7-8 can be derived from the working modes 2 and 6 by a rotation of 120° and 240°, respectively. Therefore, only the working modes 1 and 2 are studied.

### 3.6 Isoconditioning Loci

Moreover, for each Jacobian matrix and for all the poses of the end-effector, we calculate the optimum conditioning according to the orientation of the end-effector. At the other end of the spectrum, the minimum conditioning is always associated with a singular configuration.

To calculate the condition number of matrix $\overline{\mathbf{A}}$, we need the product $\overline{\mathbf{A}}\,\overline{\mathbf{A}}^T$. Figure 6 depicts the isoconditioning loci of this matrix.

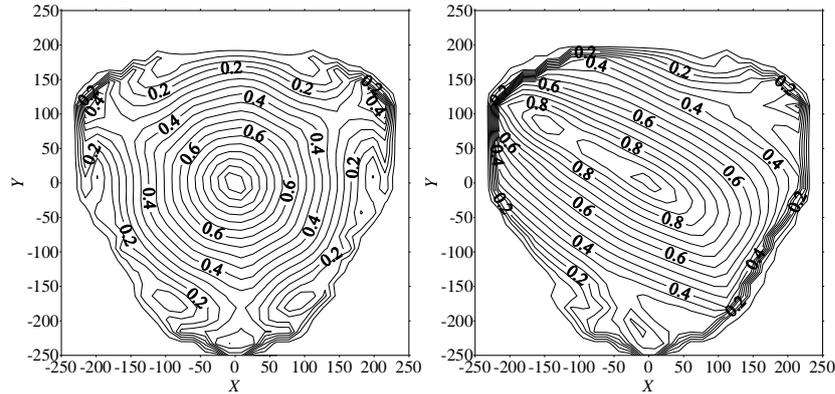

*Figure 6: Isoconditioning loci of the matrix $\overline{\mathbf{A}}$ with r=100mm, R=200mm and l=200mm*

By virtue of the diagonal form of matrix **B**, its singular values $\beta_1$, $\beta_2$ and $\beta_3$ are simply the absolute values of its diagonal entries, namely,

$\beta_1 = |m_1|$, $\beta_2 = |m_2|$ and $\beta_3 = |m_3|$

The condition number $\kappa$ of matrix **B** is thus

$$\kappa(\mathbf{B}) = \sqrt{\frac{\beta_{max}}{\beta_{min}}}$$

We depict in Fig. 7, the isoconditioning loci of matrix **B**. We notice that the loci of both working modes are identical. This is due to the absence of joint limits on the actuated joints. For one configuration, only the signs of $m_i$ change from a working mode to an other, and the condition number $\kappa$ is computed a the absolute values of $m_i$.

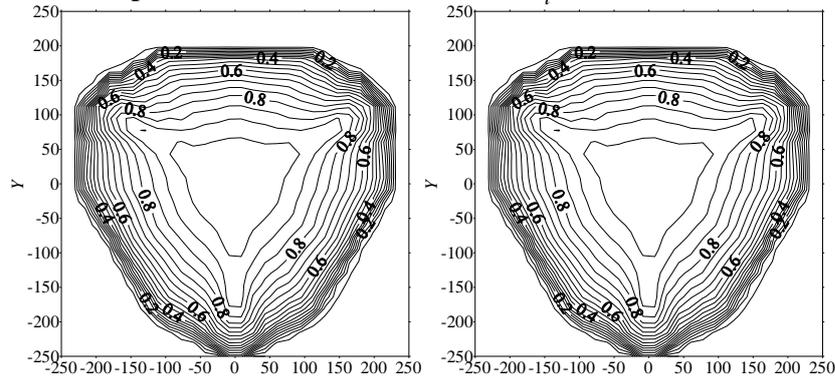

*Figure 7: Isoconditioning loci of the matrix **B** with r=100mm, R=200mm and l=200mm*

The shapes of the isoconditioning loci of $\bar{\mathbf{K}}$ (Fig. 8) are similar to those of the isconditioning loci of $\bar{\mathbf{A}}$; only the numerical values change. We can thus conclude that in our example, the singularities due to the matrix **B** are less important than those due to matrix $\bar{\mathbf{A}}$

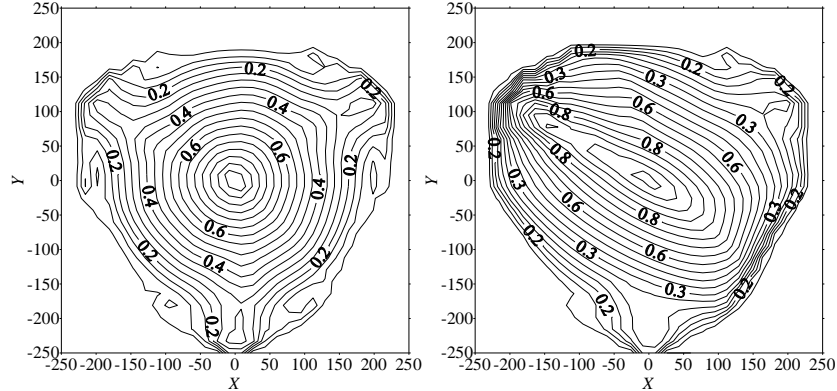

*Figure 8: Isoconditioning loci of the matrix $\bar{\mathbf{K}}$ with r=100mm, R=200mm and l=200mm*

The average of κ can be accepted as a global performance index. So, this index can allow us to compare the working modes (Table 1).

| Working modes | κ($\bar{\mathbf{A}}$) | κ(**B**) | κ($\bar{\mathbf{K}}$) |
|---|---|---|---|
| 1 | 0.354 | 0.709 | 0.329 |
| 2 | 0.511 | 0.709 | 0.460 |

*Table 1: The average value of the condition number in the Cartesian workspace*

For the first working mode, the condition number decreases regularly around the isotropic configuration. The isoconditioning loci resemble concentric circles. However, for the second working mode, the isoconditioning loci resemble elliptical curves. The average of κ is larger in the second working mode. Therefore, the prescribed trajectories should lie in an rectangular zone rather than in a square zone.

## 4  Conclusions

We produced the isoconditioning loci of the Jacobian matrices of a three-PRR parallel manipulator. This method being general, it can be applied to any three-dof planar parallel manipulator. To solve the problem of nonhomogeneity of the Jacobian matrix, we used the notion of characteristic length. This length was defined for the isotropic configuration of the manipulator. The isoconditioning curves thus obtained characterize, for every posture of the manipulator, the optimum conditioning for all possible orientations of the end-effector. This work can be used to choose the working mode which is best suited to prescribed trajectories or as a global performance index when we study the optimum design of this kind of manipulators.